# Deep Gait Tracking With Inertial Measurement Unit

Jien De Sui, and Tian Sheuan Chang

*Institute of Electronics, National Chiao Tung University, Hsinchu 300, Taiwan*

**Abstract**— This paper presents a convolutional neural network based foot motion tracking with only six-axis Inertial-Measurement-Unit (IMU) sensor data. The presented approach can adapt to various walking conditions by adopting differential and window based input. The training data are further augmented by sliding and random window samplings on IMU sensor data to increase data diversity for better performance. The proposed approach fuses predictions of three dimensional output into one model. The proposed fused model can achieve average error of 2.30±2.23 cm in X-axis, 0.91±0.95 cm in Y-axis and 0.58±0.52 cm in Z-axis.

**Index Terms**— IMU sensor, Convolutional neural networks (CNN), gait analysis, gait parameter.

## I. INTRODUCTION

Foot motion tracking is getting popular due to its value in a wide range of applications such as health care or sport training [1-2]. The gait trajectory tracking enables users to calculate various gait parameters such as stride length, stride height and stride width [3-4]. The optical based tracking methods can provide accurate tracking results but suffer from high product cost, complex setup and limited test environments [5-8]. On the other hand, the IMU (inertial measurement unit) based sensory systems allow measurement entity to be neither constrained in motion nor to any specific environments. Existing IMU based designs assumes zero velocity update and use double integral on IMU data to recreate the trajectory [9-10], which will need per sensor based or personalized calibrations to get correct results.

This paper presents the first trial of foot motion tracking with convolutional neural network (CNN). CNN is a popular deep learning approach in recent years due to its excellent performance for speech and computer vision (e.g., image classification, action or gesture recognition). CNN uses convolutional layers for feature extraction, and fully connected layers for classification or regression. Readers can refer [11] for more details. Deep learning approaches can automate feature learning and final task learning without time consuming manual processing. Several works have applied CNN on IMU data for activity recognition [12-14]. All these works are for classification. On the other hand, the proposed work is for regression that predicts foot trajectory with CNN.

In this paper, the foot trajectory per step is learned in an end-to-end way without explicit calibrations [9] or assumptions of zero velocity update [10]. The proposed deep learning model fuses three dimension predictions into one to reduce model size while improve prediction accuracy. This model also uses differential input and output to avoid absolute coordinate reference, and two data augmentations to increase data diversity for better robustness. The presented approach adopts a window based input to adapt to different length per step.

The rest of the paper is organized as follows. Section II shows the presented approach. Section III shows the results and Section IV concludes this paper.

## II. METHODOLOGY

### A. Sensor

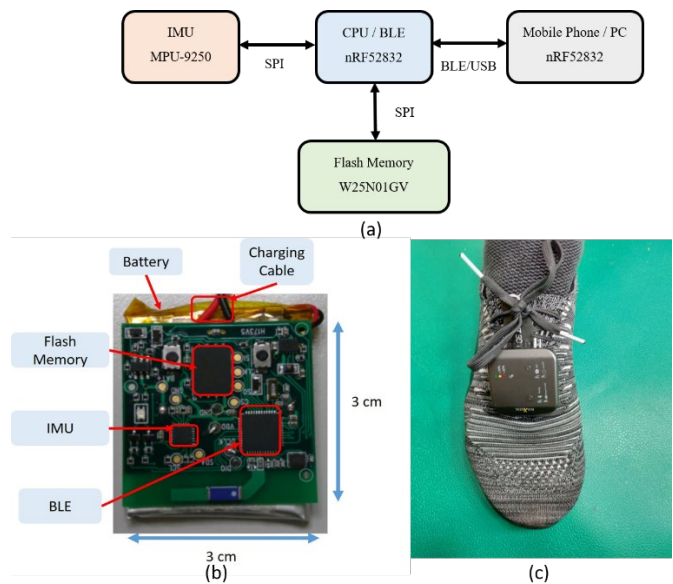

Fig. 1 The sensor system architecture (a) and its PCB (b). (c) The position of sensor on the foot.

Fig. 1 shows the sensor system architecture, which consists of a 9-axis IMU from InvenSense (MPU-9250), 1Gb flash memory for storage, and a Bluetooth chip for wireless data transmission. The IMU consists of tri-axis accelerometer, tri-axis gyroscope and tri-axis magnetometer. This system is integrated in a PCB of form factor 30

× 30 mm². This prototype consumes 9.3mA current for the highest sampling rate at 1000 Hz, and can ensure all day long continuous data recording with the 450 mAh rechargeable 3.7 V Li-ion battery. The recorded data are transmitting to mobile phone or PC for further data processing.

The sensor is mounted and fixed on the shoe as shown in Fig. 1. The sampling rate of this measurement is set to 500 Hz. The accelerometer has a dynamic range of ± 16 g and the gyroscope has a dynamic range of ± 2000 °/s [15].

### B. Data Collection

We collect the walking test of 10 normal people aged from 20 to 30 as our dataset. The walking test is free walking that includes either a straight line walk, or circular walk. The walking speed is ranged from 0.45m/s to 1.75m/s (1.64km/h to 6.30km/h) to test slower, preferred and faster walking speed. The ground truth data is collected by a motion capture system, the VICON NEXUS camera system. The average error of the VICON camera system is 0.1 to 0.3mm based on the environment setting and the condition of cameras. The sampling rate of VICON system is set to 100 Hz.

### C. Data augmentation

Before data augmentation, the amount of original dataset is 90,255 sample points (270 seconds). To fit the fixed size input requirement of the CNN model, we use a window based input that consists of 150 sample points per window as shown in Fig. 2, which can help adapt to different length per step. To further increase training data amounts, we implement two data augmentation methods for the proposed CNN model: sliding window and random window sampling. The first method, sliding window, is to split one step of sensor data to 150-sample-point windows with 10-sample-point overlapping between windows, as shown in Fig. 2. Similar operations are also applied to the ground truth but use 30-sample-point windows with 2-sample-point overlapping since the sampling rate of the Vicon system is roughly five times smaller than that in IMUs. With this method, the available data amount will become 110,856 sample points (744 seconds), which is 2.8 times than the original one.

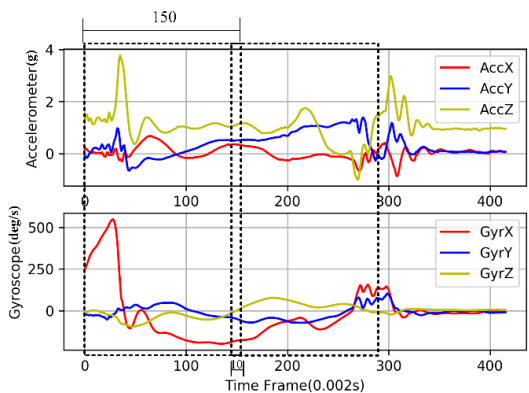

Fig. 2 Data augmentation with the window based method.

The second data augmentation, random window, randomly chose five 150-sample-point windows from each step instead of the fixed and overlapped approach as the sliding window. With this method, the data amount will become 344,935 sample points (2315 seconds), which is 8.6 times than the original one.

In this paper, we will combine the datasets generated by these two data augmentations for model training. The amount of whole dataset will become 455,791 sample points (3059 seconds), which is 11.3 times than the original one. Note that the testing data is also split to 150-sample-point windows but without any overlapping.

### D. Data preprocessing

The collected data are first segmented into steps. One walking step length is roughly fewer than 400 sampling points as shown in Fig. 2. We use 80% of steps for training, 10% of steps for validation and 10% of steps for testing. The proposed motion tracking is then applied for each step.

Beyond segmentation, directly applying raw data to the regression model for motion trajectory faces the difficulty to track absolute coordinates of the ground truth data, which is not reasonable since similar trajectory could occur just with different baseline. To avoid such problem, we use the differential value between consecutive samples, as our training and testing data. Thus, a 150-sample-point sensor data will become 149 sample points, and 30-sample-point ground truth will become 29 sample points. Such differential value approach can make the model more robust.

Thus, input sensor data become 6 × 149 ×1 (sensors × time segment size × channel), where 6 for 6-axis sensor data (tri-axis accelerometer, and tri-axis gyroscope) and 149 for data length after preprocessing. The ground truth data is 3 × 29, where 3 for three-dimension space axis.

### E. Network Architecture

Fig. 3 shows the two proposed network architectures based on convolutional layers and fully connected layers. The first one uses a single model to predict (X, Y, Z) positions at the same time, denoted as the *fused model*. The second one uses three separated models to predict (X, Y, Z) positions respectively, denoted as the *independent model*. The fused model has shared network layers for (X, Y, Z) predictions, while the independent model can optimize the network tailored for each axis. All these models use the same input sensor data structure and the same network architecture, which includes nine convolutional layers and two fully connected layers.

Table 1 shows the detailed network architectures. For the convolutional layers we set channel number N1 = 64, N2 = 64, N3 = 128, N4 = 128, N5 = 256, N6 = 256, N7 = 512, N8 = 512 and N9 = 1024, with 3×3 filter size and Rectified Linear Unit (ReLU) as the activation function. In addition, batch-normalization and max pooling operations are applied after activation function successively. For max pooling, the filter size is set to 1 × 2 (sensors × time, downsampling only in the time axis) so that we can get wider receptive field along the time axis instead of sensor axis to avoid information loss.

In fully connected layers, we set N10 = 1024 and N11 = 512 with ReLU as the activation function. In addition, we apply dropout here to prevent overfitting with a dropout probability, p = 0.5, for each fully connected layer. For the final output layers, a fully connected layer is used to generate the final tracking data. For the fused model, we will have three output layers for (X, Y, Z) respectively. For the independent model, each axis will have its own model and output.

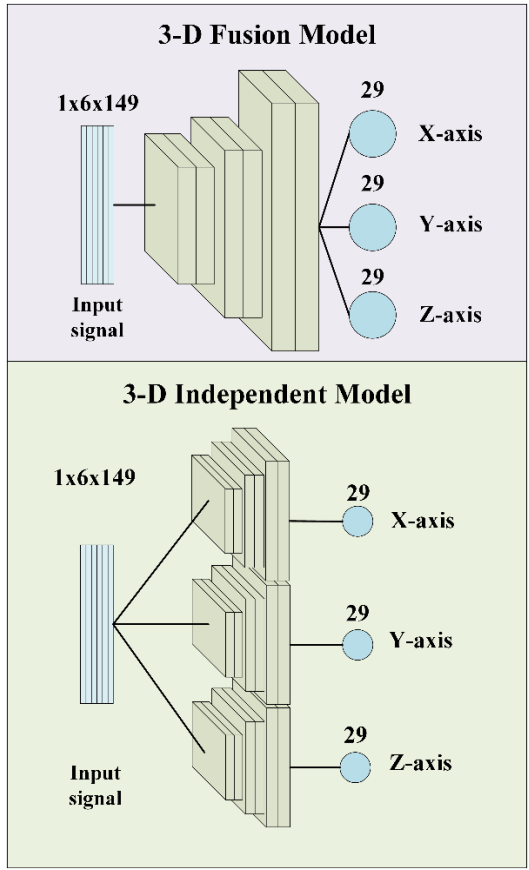

Fig. 3 Two different architectures to predict 3-D outputs

Table 1. The detail of both 3-D Fusion model and 3-D Independent model. Including two architectures.

| Number | 9 Layer Conv | | 5 Layer Conv | |
|---|---|---|---|---|
| | Layers | Channel | Layers | Channel |
| 1 | Conv | 64 | Conv | 64 |
| | BN | | BN + MaxPooling | |
| 2 | Conv | 64 | Conv | 128 |
| | BN | | BN + MaxPooling | |
| 3 | Conv | 128 | Conv | 256 |
| | BN + MaxPooling | | BN + MaxPooling | |
| 4 | Conv | 128 | Conv | 512 |
| | BN + MaxPooling | | BN + MaxPooling | |
| 5 | Conv | 256 | Conv | 1024 |
| | BN + MaxPooling | | BN + MaxPooling | |
| 6 | Conv | 256 | | |
| | BN + MaxPooling | | | |
| 7 | Conv | 512 | | |
| | BN + MaxPooling | | | |
| 8 | Conv | 512 | | |
| | BN + MaxPooling | | | |
| 9 | Conv | 1024 | | |
| | BN + MaxPooling | | | |
| | Flatten | | Flatten | |
| 10 | Dense+Drop | 1024 | Dense+Drop | 1024 |
| 11 | Dense+Drop | 512 | Dense+Drop | 512 |
| 12 | Output | 29+29+29 | Output | 29 |
| Parameter | 16,274,711 | | 17,069,015 | |

## III. RESULT

### A. Training

The loss function defines the error between predicted output and ground truth. For the independent model, we use Root-Mean-Square-Error (RMSE) as the loss function for each model. For the fused model, we combine all the RMSE loss values from each axis with weighted factors to balance between different scales of three axis values.

$$Loss = \mathrm{RMSE}(X_{axis}) + w_1 \times \mathrm{RMSE}(Y_{axis}) + w_2 \times \mathrm{RMSE}(Z_{axis})$$

In this paper, we set the $w_1$=10 and $w_2$ = 10. We use random subsets of the training dataset (mini-batches), and train the model for 1000 epoches with batch size 100. For optimization, we use the Adam optimizer with default setting of β = (0.9,0.999) and ϵ = $1e^{-8}$, set learning rate to 0.01, and initialize model parameters with random assignments from the normal distribution.

The model was implemented with Pytorch, running on Geforce GTX 1080 GPU. For the output, one walking step contains two or three data windows. Thus, we will concatenate these outputs to rebuild motion trajectory.

### B. Result

Fig. 4 shows the rebuilt trajectory, which is one of testing results in 2D and 3D plots. The proposed model can effectively track the gait motion based on IMU sensor data.

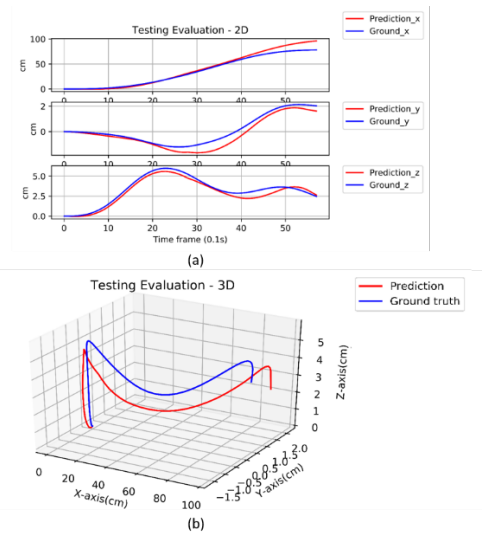

Fig. 4 (a) 2D tracking results (b) 3D tracking results

Table 2 shows the error analysis of the fused and independent approaches with five or nine convolutional layers. We also test the effectiveness of proposed data augmentation methods. Based on the results, we have the following observations.

Table 2 shows that the fused method has lower errors than the independent method for almost every kind of cases. The 3-D fusion model can be regarded as multi-task learning. With such multi-task networks, we will not only excel at accuracy due to regularization by multi-task learning to reduce the probability of overfitting [16], but also reduce model complexity to one model instead of three models.

The result also shows that the deeper network has lower error since it can extract better features.

Table 2 also shows the effectiveness of the two data augmentation methods. When compared to original design without window based input, window based data augmentations can improve the errors significantly. In which, sliding window has lower accuracy than the random window, which could be due to similar window partitioning between training and testing data in this case. Furthermore, combining the two methods can achieve the lowest errors due to larger training dataset. The differential input approach is necessary for model success, as shown in Table 2. The errors of raw data are much higher than those with differential data.

In summary, the proposed approach with 3D fusion network and combined data augmentations can achieve average error of 2.30 cm in X-axis, 0.91 cm in Y-axis and 0.58 cm in Z-axis.

Above simulations mix steps from all walkers with random step selection for training/validation/testing. For validation purpose, we have also applied a 6-fold cross validation with independent walkers on training set and testing set. In this scenario, 10% of steps from the training set will be the validation set. The lowest error of the independent walker test is 7.78±5.31cm, 1.49±1.14cm, 1.03±0.53cm for X, Y, and Z respectively, which is also from the 3D fusion with 9 convolution layers. This error is higher than the mixed walker test. To close the gap, more training data is desired.

Table 2. The result of proposed models on different conditions

| Architecture | Dataset | X error(cm) | Y error(cm) | Z error(cm) |
|---|---|---|---|---|
| 3-D Fusion (5 layer Conv) | w/o window | 6.56±5.91 | 1.83±1.31 | 0.34±0.15 |
| | sliding | 4.52±3.90 | 1.43±1.22 | 0.89±0.83 |
| | random | 5.26±4.43 | 1.34±1.32 | 0.89±0.63 |
| | random + slide | 2.76±2.98 | 0.80±0.77 | 0.63±0.55 |
| 3-D Separated (5 layer Conv) | w/o window | 6.58±4.57 | 2.39±1.71 | 0.51±0.30 |
| | Sliding | 3.11±2.27 | 1.13±0.85 | 0.90±0.64 |
| | Random | 6.34±4.87 | 1.59±1.24 | 1.14±1.06 |
| | random + slide | 3.30±4.90 | 0.99±0.79 | 0.72±0.38 |
| 3-D Fusion (9 layer Conv) | w/o window | 3.85±3.18 | 1.64±1.06 | 0.32±0.18 |
| | sliding w/o diff | 6.26±5.67 | 1.80±1.87 | 1.15±1.17 |
| | sliding | 2.79±2.30 | 0.96±0.82 | 0.73±0.72 |
| | random | 5.80±5.57 | 1.44±1.23 | 0.99±0.95 |
| | **random + slide** | **2.30±2.23** | **0.91±0.95** | **0.58±0.52** |
| 3-D Separated (9 layer Conv) | w/o window | 6.05±5.34 | 1.91±1.68 | 0.42±0.18 |
| | sliding | 2.46±2.36 | 0.99±0.88 | 0.69±0.65 |
| | random | 5.92±5.93 | 1.43±1.37 | 1.26±1.09 |
| | random + slide | 2.71±3.00 | 0.79±0.76 | 0.65±0.55 |

## IV. CONCLUSION

This paper has presented a deep gait tracking method on IMU sensor data with convolutional neural network. The proposed approach uses differential and window based input to successfully adapt to different walking conditions. Model performance are further improved with data augmentation methods, sliding and random window samplings. The final 3D fused model enables lower complexity while achieves average error of 2.30 cm in X-axis, 0.91 cm in Y-axis and 0.58 cm in Z-axis.